\documentclass{article}
\usepackage{authblk}
\usepackage{spconf,amsmath,epsfig}

\let\OLDthebibliography\thebibliography
\renewcommand\thebibliography[1]{
  \OLDthebibliography{#1}
  \setlength{\parskip}{0pt}
  \setlength{\itemsep}{0pt plus 0.3ex}
}

\usepackage{etoolbox}
\usepackage{bm}
\usepackage{amsthm}%
\usepackage{subfigure}%

\newcommand{\MyMapTemplatePrefixc}[4]{\expandafter#1\csname#3#4\endcsname{#2{#4}}} % it remembles a template: \#3#4 --> #2{#4}
\forcsvlist{\MyMapTemplatePrefixc {\def} {\mathcal}{c}} {A,B,C,D,E,F,G,H,I,J,K,L,M,N,O,P,Q,R,S,T,U,V,W,X,Y,Z}  % e.g., \cA --> \mathcal{A}

\newcommand{\MyMapTemplatePrefixtb}[5]{\expandafter#1\csname#4#5\endcsname{#2{#3{#5}}}} % it remembles a template: \#3#4 --> #2{#4}
\forcsvlist{\MyMapTemplatePrefixtb {\def} {\tilde}{\mathbf}{t}} {A,B,C,D,E,F,G,H,I,J,K,L,M,N,O,P,Q,R,S,T,U,V,W,X,Y,Z}  % e.g., \tA --> \tilde{A}
\forcsvlist{\MyMapTemplatePrefixtb {\def} {\tilde}{\mathbf}{t}} {0,1,a,b,c,d,e,f,g,h,i,j,k,l,m,n,o,p,q,r,s,u,v,w,x,y,z}  % e.g., \ta --> \tilde{a}

\newcommand{\MyMapTemplateNoPrefix}[3]{\expandafter#1\csname#3\endcsname{#2{#3}}}
\forcsvlist{\MyMapTemplateNoPrefix {\def} {\mathbf} } {0,1,a,b,c,d,e, f, g, h, i, j, k, l, m, n, o, p, q, r, u, v, w, x, y, z} % e.g., \a --> \mathbf{a}
\forcsvlist{\MyMapTemplateNoPrefix {\def} {\mathbf} } {A,B,C,D,E,F,G,H,I,J,K,L,M,N,O,P,Q,R,S,T,U,V,W,X,Y,Z}  % e.g., \A --> \mathcal{A}

\def\etal{\emph{et al.}}

\usepackage{booktabs} % Nice tables
\usepackage[table,dvipsnames]{xcolor}
\definecolor{rowblue}{RGB}{220,230,240}

\usepackage{fancyhdr}
\fancypagestyle{FirstPage}{

\rfoot{978-1-6654-3864-3/21/\$31.00 \copyright2021 IEEE}

}
\begin{document}\sloppy

% Example definitions.
% --------------------
\def\x{{\mathbf x}}
\def\L{{\cal L}}

% Title.
% ------
\title{Improving Facial Attribute Recognition by Group and Graph Learning}
%
% Single address.
% ---------------
% \name{}
% \address{Paper ID: 102}
\name{\vspace{-2mm}Zhenghao Chen$^\dag$, Shuhang Gu$^\dag$, Feng Zhu$^\ddag$, Jing Xu$^*$ and Rui Zhao$^\ddag$}

%Address and e-mail should NOT be added in the submission paper. They should be present only in the camera ready paper. 
\vspace{-2mm} 
\address{$^\dag$\textit{The University of Sydney},
          $^\ddag$\textit{Sensetime Group Limited and} 
          $^*$\textit{Applied Research Center, Tencent PCG} \\
         \small{zhenghao.chen@sydney.edu.au, shuhanggu@gmail.com, \{zhufeng, zhaorui\}@sensetime.com and
          eudoraxu@tencent.com}}

\maketitle

\begin{abstract}
Exploiting the relationships between attributes is a key challenge for improving multiple facial attribute recognition. In this work, we are concerned with two types of correlations that are spatial and non-spatial relationships. For the spatial correlation, we aggregate attributes with spatial similarity into a part-based group and then introduce a Group Attention Learning to generate the group attention and the part-based group feature. On the other hand, to discover the non-spatial relationship, we model a group-based Graph Correlation Learning to explore affinities of predefined part-based groups. We utilize such affinity information to control the communication between all groups and then refine the learned group features. Overall, we propose a unified network called Multi-scale Group and Graph Network. It incorporates these two newly proposed learning strategies and produces coarse-to-fine graph-based group features for improving facial attribute recognition. Comprehensive experiments demonstrate that our approach outperforms the state-of-the-art methods.%
\end{abstract}
\vspace{-2mm}
\section{Introduction}
\thispagestyle{FirstPage}
\vspace{-2mm}
\label{sec:intro}
% Facial attributes are used to represent visual appearances on a face with human-understandable descriptions~\cite{liu2015deep}. 
Interpreting facial attributes plays an important role in many multimedia applications (\textit{e.g.,} face-based augmented reality).
Most researchers treat facial attribute recognition as a Multi-Task Learning (MTL) problem as it normally requires to predict multiple facial attributes at once. Hence, better exploring and incorporating the correlations between different attributes can efficiently improve the recognition performance~\cite{hand2017attributes}.
%
%While recently, based on the observation that certain facial region naturally corresponds to a combination of attributes \cite{liu2015deep, facesurvey}, a sizable body of works \cite{hand2017attributes, cao2018partially, han2018heterogeneous} has provided evidence that spatial similarity is beneficial for recognition. 

The spatial similarity is the most straightforward relationship as a certain facial region naturally corresponds to a combination of attributes~\cite{liu2015deep}. 
A sizable body of works \cite{hand2017attributes, cao2018partially, han2018heterogeneous} have already indicated that using spatial similarity can benefit the recognition. 
However, existing methods simply exploited spatial similarity by extracting the shared feature from spatial-related attributes, thus had limited capacity in profiting from spatial correlation information.
%
%Inspired by some approaches capture facial part-based information to improve individual attribute estimation by auxiliary data \cite{zhang2014panda, rudd2016moon, kalayeh2017improving, he2018harnessing} or weakly-supervised learning \cite{ding2018deep,li2018landmark}. 
%
To better leverage the spatial correlation, 
we develop a method named Group Attention Learning (GAL) to apply part-based learning and spatial correlation together. 
We not only aggregate correlated attributes for learning shared feature extractors
but also produce part-based group attention masks.
In this way, we further leverage the spatial knowledge to learn group-specific attentions, which greatly strengthens shared part-based group features.
%Attributes located in nearby positions will be assigned into one part-based group and jointly learn a shared part-based attention mask through a Group Attention Network. 
%Such attention will then strengthen the shared group feature. The performance of part-based learning will also be enhanced by taking advantage of spatial similarity to reduce the complexity and difficulty for learning the attention.

% Besides further exploiting spatial correlation to enhance intra-group feature learning, we also propose a group-based Graph Correlation Learning strategy to leverage non-spatial relationship between facial attributes.
Besides further leveraging spatial correlation to enhance intra-group feature learning, we also exploit
non-spatial correlation. It refers to the fact that attributes on different facial regions may still related to each other.
For instance, people having the lipstick (on the mouth) are more likely to wear the necklace (on the neck)~\cite{hand2017attributes}.
%
% Compared with spatial correlation, such non-spatial correlation is much harder to use.
%
In order to incorporate such correlation hints, 
% adaptive learning \cite{lu2017fully, he2017adaptively} and Neural Architecture Search (NAS) \cite{huang2018gnas} 
some adaptive learning methods~\cite{lu2017fully,huang2018gnas}
have been recently deployed for mining the relationships between pairs of attributes on different regions.
Despite improving
% the overall recognition 
holistic recognition performance, the large search space~\cite{lu2017fully} between pairs of attributes inevitably brings a large computation burden in the training.
%of network. 
%
Moreover, all existing approaches assumed that the relationships between different attributes are undirected.
%, e.g.   
% (\textit{i.e.,} attributes in a pair have the same influence to each other).
%
Obviously, the mutual influence between attributes could be different, ignoring the direction of influence greatly affects the exploitation of correlations.
% between attributes. %
%
To address the above issues, we develop a Graph Correlation Learning (GCL) upon our part-based groups. 
Concretely, a directed group-based Graph Neural Network (GNN) is conducted to explore non-spatial inter-group affinity. % from groups.
Such affinity is then used to update part-based group features to graph-based group features, which will contain both spatial and non-spatial information. 
As a result, our strategy is capable of mining directional non-spatial correlation messages between heterogeneous groups for improving recognition performance.

%Besides the spatial correlation, attributes in different facial regions could also possibly be related.
% This is so-called non-spatial correlation.
%However, such a connection is much harder to mine and use as it is not as obvious as a spatial similarity. 
%Some recent methods directly employ well-known algorithms such as adaptive learning \cite{lu2017fully, he2017adaptively} and Neural Architecture Search (NAS) \cite{huang2018gnas} to exploit the relationship for every pair of individual attributes. This way correspondingly brings a large search space \cite{lu2017fully} and normally ignores the direction of influence \cite{hand2017attributes, han2018heterogeneous}. 
%Different from those methods that only discover the relationship for individual attributes, we develop a new strategy called Graph Correlation Learning upon our part-based groups. 
%A directed group-based graph network is conducted to explore non-spatial inter-group affinity from heterogeneous groups. Such affinity is then used to consolidate learned part-based group features. Eventually, each group will have both its spatial feature obtained by spatial similarity of its attributes and non-spatial correlation messages from other groups.

Overall, we propose a unified framework Multi-scale Group and Graph Network (MGG-Net) to integrate these two novel learning strategies on multi-scale feature,
in which we outperform state-of-the-art approaches on prediction and balanced accuracy.
%
%MGG-Net improves both prediction \cite{liu2015deep} and balanced \cite{huang2019deep} accuracy, while most approaches only target one measurement.
%
% We achieve \textbf{$92.00\%$} prediction accuracy and \textbf{$89.19\%$} balanced accuracy on CeleBA dataset and  \textbf{$87.20\%$} accuracy on LFWA dataset
%
Our contributions are listed as followed:
\vspace{-2mm}
\begin{itemize}
 % \item A novel method Local Part-based Group better learns and applies the part-based feature for attribute estimation.
  \item A group attention mechanism GAL leverages intra-group spatial correlation of facial attributes and part-based learning to produce part-based group features.
  \item A group-based graph learning strategy GCL exploits the inter-group non-spatial correlation between part-based attribute groups to refine learned group features. 
  \item A unified framework MGG-Net applies our approach to multi-scale feature levels and significantly improves the recognition performance on CeleBA and LFWA.
  % all AAAI/IJCAI papers regarding attributes include performance a one of contribution
\end{itemize}

\section{Related Work}
\vspace{-2mm}
\textbf{Facial Attribute Recognition.} 
To estimate a single facial attribute, traditional methods mostly used hand-crafted features along with a classifier~\cite{kumar2008facetracer}. The first deep learning facial attribute prediction method was proposed by Liu~\etal~\cite{liu2015deep}. Following that, various Convolution Neural Network (CNN) methods were proposed~\cite{zhong2016face, rudd2016moon} to address this task. 
Part-based learning is the most well-known strategy for single attribute prediction. Zhang~\etal~\cite{zhang2014panda} employed the facial landmarks to attain defined regions and then embedded them into Poselet. Kalayeh~\etal~\cite{kalayeh2017improving} obtained segmentation masks from another face parsing domain and fused such masks to original features. He~\etal~\cite{he2018harnessing} learned abstract facial images and leveraged them as additional part-based features. 
Rather than relying on auxiliary data to obtain localization cues, some recent methods attempted to use weakly-supervised approaches to generate attentions~\cite{ding2018deep,li2018landmark,ji2020unsupervised}. For example, Ding~\etal~\cite{ding2018deep} used a localization network to learn the attention for every individual attribute.
% Li~\etal~\cite{li2018landmark} extracted part-based features with a set of comprehensive transformation sub-networks.
% Attention methods are also widely applied to pedestrian attribute prediction \cite{sarafianos2018deep} as pedestrian attributes can benefit from very rough part-based features such as upper, middle and lower body which are trivial to learn. 
% However, learning attentions for facial attributes is more complicated. 
% For instance, the attribute "Makeup" requires large and coarse spatial features for recognition, while we need fine-grained attention to detect the attribute "Earrings". 
In our method, we employ part-based groups to learn part-based attentions. Our design is fully based on the spatial similarity of attributes. Such affinity reduces the difficulty and complexity of attention learning. Moreover, a coarse-to-fine multi-scale architecture is employed in our network, so that different resolutions of part-based features can benefit various types of attributes.

\textbf{Multi-Task Learning.} 
MTL is a learning paradigm that jointly learns multiple tasks. Its main challenge is how to exploit relationships across different tasks so that positive correlation information can then be used to enhance holistic performance. For multiple facial attributes recognition, some MTL-based approaches aggregated attributes into groups to extract the shared feature~\cite{hand2017attributes, cao2018partially, han2018heterogeneous,gong2020jointly,yang2020hierarchical}. Those works mostly designed groups according to prior spatial knowledge and only focused on the affinity inside the designed group. Differently, our method not only generates a better shared spatial feature using the group attention but also exploits the cross-group communication by a graph. Other methods used learning-based approaches to obtain the correlation. Lu~\etal~\cite{lu2017fully} adaptively learned a thin-to-wide network, while Huang~\etal~\cite{huang2018gnas} employed a greedy search. Though those methods could intuitively discover non-spatial correlations, they had to explore through all attribute pairs, which causes a large search space and complexity. 
Differently, our method explores the graph-based non-spatial affinity through part-based groups rather than individual attributes to avoid this issue.
% Our method can avoid this issue by taking groups instead of attributes to reduce the number of nodes in the graph learning.

% with higher resolution in shallower layers

\textbf{Graph Neural Network.} 
GNN is proposed to deal with the non-linear data formed with graph structure. Among various GNN-based methods,
our work is mostly inspired by Graph Attention Network (GAT)~\cite{velivckovic2017graph}, which estimated an adjacency matrix through a multi-head attention-based mechanism. Recently, some graph-based approaches were also introduced to the pedestrian attribute prediction task~\cite{li2019visual}, which learned the relationships between individual attributes.
In this work, rather than building the graph upon individual attributes, we construct a graph by taking part-based groups as nodes. In this way, we can model our exploration study to estimate inter-group non-spatial affinities from groups to groups. A two-way directed GNN is then conducted to learn bilateral relationships between part-based groups. Therefore, for each group, messages it sends to others will be distinguished from the messages it receives. Lastly, different from works \cite{li2019visual} in pedestrian attribute estimation that used complicated multi-stage ensemble model for node processing and graph modeling, we incorporate group feature learning together with graph-based correlation learning in a unified model. So we can optimize it in an end-to-end manner. 

\begin{figure*}[h]
\begin{center}
\includegraphics[width=\linewidth]{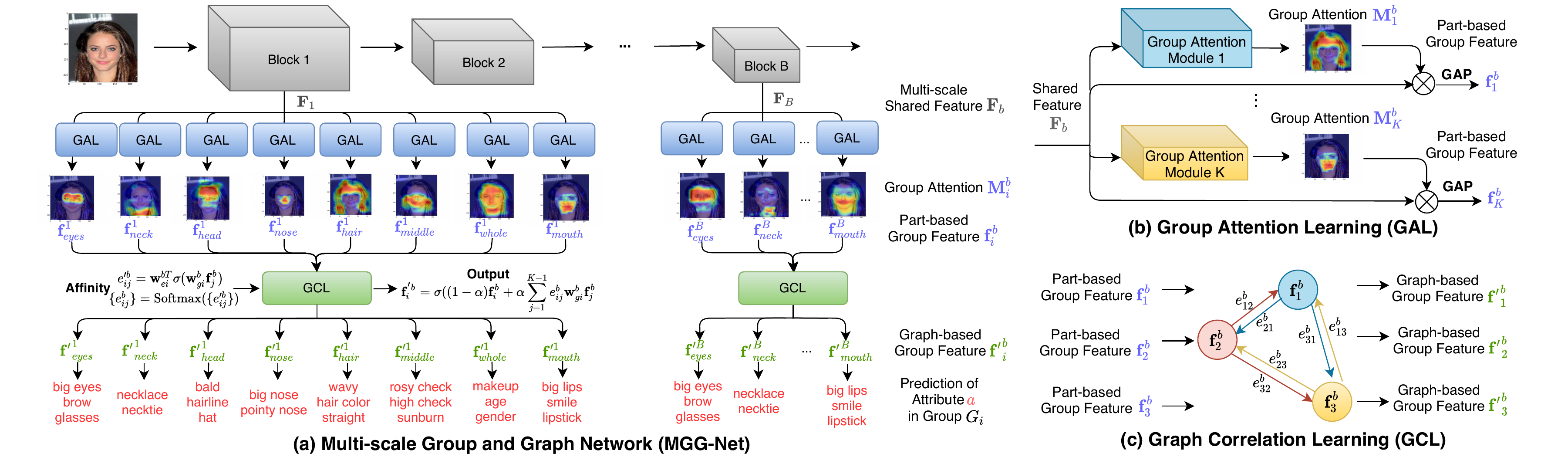}
\caption{An overview of our MGG-Net (a), GAL (b) and GCL (c). We assign multiple attributes $a$ to different part-based groups $\mathcal{G}_i$ according to their spatial similarity. 
The backbone first extracts multi-scale shared feature $\F_b$ from the $b$th block. Then, GAL will generate group attention mask $\M^b_i$ for each $\mathcal{G}_i$ from $\F_b$. Such mask $\M^b_i$ will multiply with $\F_b$ and produce the part-based group feature $\f_i^b$ by global average pooling (GAP). After that, GCL will update part-based group feature $\f^b_{i}$ to graph-based group feature $\f'^b_i$ by using affinities between part-based groups. Eventually, $\f'^b_{i}$ will contain both group-based spatial and graph-based non-spatial correlation information. It would then be used to predict all facial attributes $a$ in part-based group $\mathcal{G}_i$.}
\label{fig:network}
\end{center}
\end{figure*}

\section{Methodology}
\vspace{-1mm}
%  In this section, we will first introduce our baseline Multi-Task Convolution Neural Network (MT-CNN), followed by our main methods Group Attention Learning and Graph Correlation Learning. In the end, we will show two objective losses for prediction and balanced accuracy.

\subsection{Preliminaries}
\vspace{-1mm}

\textbf{Problem Formulation.}
Given a facial image with $N$ pre-defined attributes $a$, our goal is to predict the probability $Y_a$ for each attribute $a$, that achieves the minimal average prediction error $\frac{1}{N} \sum_{a=1}^N || Y_a, \hat{Y}_a||$, where $\hat{Y}_a$ is the attribute label.

\textbf{Part-based Group.}
We split $N$ facial attributes into $K$ part-based groups according to their location similarity.
%Specifically in our implementation, we design our groups as Table.~\ref{tab:group} ($N=40$ and $K=8$).
We define such a group as $\mathcal{G}_i$, which represents the group for facial part $i$. $a \in \mathcal{G}_i$ indicates that attribute $a$ is located at the part $i$ and belongs to group $\mathcal{G}_i$. We then apply our approach to obtain the feature for each group $\mathcal{G}_i$. Such feature will be used to predict the probability $Y_a$ for attribute $a$, \textit{s.t.,} $a \in \mathcal{G}_i$. 

% Our work is conducted on a shared-split Multi-Task CNN,
% which uses a CNN network to extract the shared feature for multiple learning tasks. In the facial attribute recognition problem, the prediction of an attribute is considered as one task. While our method will adopt such shared features $F_b$ on different blocks $b$ to further apply the Group and Graph Learning, for our baseline, we will directly use the output $F$ of the last convolution layer with a Global Average Pooling (GAP) as the shared feature to predict the multiple facial attributes. 
% Each prediction score $Y^{a}_{base}$ for attribute $a$ will be obtained by a binary classifier $\gamma_{a}$ with a Sigmoid activation function $\psi$ and learned parameters $\theta_{a}$.

% \begin{equation} \label{eq1-0}
% \begin{split}
% Y^{a}_{base} & = \psi(\gamma_a(F, \theta_{a}))
% % L_{base} & = \sum_{a=1}^{N} Loss(Y^{a}_{base}, Y^{\prime}_a)
% \end{split}
% \end{equation}
\subsection{Group and Graph Learning}
\vspace{-1mm}
% We propose a novel method Group and Group Learning to learn a group feature for each part-based group $\mathcal{G}_i$. 
We propose MGG-Net (Fig.~\ref{fig:network} (a)) to apply our approach and produce multi-scale group features.
Our backbone network extracts coarse-to-fine shared features through multi-scale feature blocks. In this work, we will use $B$ such blocks, where $\F_b$ denotes the shared feature extracted from the $b$th block.
Our MGG-Net will then generate group features from $\F_b$ by incorporating GAL (Fig.~\ref{fig:network} (b)) and GCL (Fig.~\ref{fig:network} (c)).

\textbf{Group Attention Learning.}
% \begin{figure}[h]
% \begin{center}
%   \includegraphics[width=0.8\linewidth]{pics/attention.pdf}
% \end{center}
%   \caption{Group Attention Network at block $b$. The shared feature $\F_b$ extracted from block $b$ is fed into $K$ group attention modules to generate $K$ attention masks $\A_i^b$. Each mask will be multiplied with $\F_{b}$ to obtain the part-based group feature $\f^b_{i}$ for group $\mathcal{G}_i$ and block $b$ by global average pooling (GAP).}
% \label{fig:groupattention}
% \end{figure}
GAL is to obtain the part-based group features for different groups by using intra-group spatial similarity and part-based learning.
On each block $b$, we use Group Attention Modules to extract $K$ group attention masks from shared feature $\F_b$. Each module contains a set of convolution, batch normalization and ReLU activation layers to produce a learned group attention mask $\M^b_{i}$ for group $G_{i}$. 
Such mask $\M^b_{i}$ will then be multiplied with shared feature $\F_{b}$ and the product will be used to generate part-based group feature $\f_{i}^b$ for group $\mathcal{G}_i$ by global average pooling (GAP).
Eventually, we will obtain $B \times K$ such part-based group features $\f_{i}^b$, where $B$ is the number of blocks and $K$ is the number of groups. The process is denoted as Eq~(\ref{eq1}), where $\phi$ represents Group Attention Modules and $\theta^b_{i}$ are learned parameters. 
% This group feature will be used to predict all scores $Y_{group}^{ba}$ for the attribute $a$ in group $\mathcal{G}_i$ (i.e., $a \in G{i}$) on block $b$. 
% In Eq.~\ref{eq2}, $\gamma_{ba}$ and $\theta_{ba}$ are the classifier and learned parameters for attribute $a$ and block $b$ while $\psi$ is Sigmoid activation function.
\begin{equation} \label{eq1}
\begin{split}
\M^{b}_i & = \phi(\F_{b}, \theta^{b}_{i}) \\
\f^{b}_i & = \text{GAP}(\F_{b} \M^{b}_{i}) \\
% Y^{ba}_{group} & = \psi(\gamma_{ba}(F_{bi}, \theta_{ba}))
\end{split}
\end{equation}

\textbf{Graph Correlation Learning.}
% \begin{figure}[!h]
% \begin{center}
%   \includegraphics[width=0.8\linewidth]{pics/gcn.pdf}
% \end{center}
%   \caption{Part-based group feature $\f_i^b$ will be fed into a two-headed directed graph $(\mathcal{F}_b,\mathcal{E}_b)$ to receive the message from other groups and obtain the graph-based group feature $\f'_i^b$.}
% \label{fig:gcn1}
% \end{figure}
While GAL manipulates the spatial correlation of attributes inside a part-based group, GCL will be used to explore the non-spatial inter-group affinity between all part-based groups to enhance the group features. 
Specifically, we build a two-way directed GNN to model part-based group features $\f^b_i$ at scale $b$. Here, we use the learned part-based group features $\f^b_i$ from GAL as nodes and the affinity values sent from $\f^b_j$ to $\f^b_i$, $e^b_{ij}$ as edges. We assume that the message sent from $\f^b_j$ to $\f^b_i$ is not identical to the message from $\f^b_i$ to $\f^b_j$ (\textit{i.e.,} $e^b_{ij} \ne e^b_{ji}$). 
Then, $\f^b_i$ will be updated by all affinity values of $e^b_{ij}$, which are received from other nodes $\f^b_j$. To compute such values, we learn a graph transfer weight $\w^b_{gi}$ and an edge weight $\w^b_{ei}$ for the node $\f^b_i$. The node $\f^b_{j}$ will be multiplied by $\w^b_{gi}$ and then activated by a ReLU activation function. After that, we multiply the transpose of $\w^b_{ei}$ to compute a value $e'^b_{ij}$. A Softmax function will then be applied for the value set $\{e'^b_{ij}\}$ to produce the affinity value set $\{e^b_{ij}\}$ with $K-1$ values.
Once we obtain all affinity values $e^b_{ij}$, the network will update part-based group feature $\f^b_i$ to graph-based group feature $\f'^b_i$ by the layer-wise graph attention mechanism. 
To this end, the graph-based group feature $\f'^b_i$ would contain both its own part-based information and correlation information from other part-based group features.
The process is denoted as Eq~(\ref{eq2}), 
where the $\sigma$ represents ReLU activation function and $\alpha$ is a constant number. 
\begin{equation} \label{eq2}
\begin{split}
e'^b_{ij} &= \w_{ei}^{bT}  \sigma(\w^b_{gi} \f^b_j ) \\
\{e^b_{ij}\} &= \text{Softmax}(\{e'^b_{ij}\}) \\
\f'^b_{i} &= \sigma((1 - \alpha)\f^b_{i} + \alpha \sum^{K-1}_{j=1} e^b_{ij} \w^b_{gi} \f^b_j)
\end{split}
\end{equation}

% On each block $b$, the graph feature $F^{\prime}_{bi}$ can be used to predict all attributes $a$ in group $\mathcal{G}_i$ (i.e., $a \in \mathcal{G}_i$) similar to Eq.~\ref{eq1-2}. $F^{\prime}_{bi}$ will be fed into $N \times B$ classifiers $\gamma_{ba}$ to predict $Y^{ba}_{graph}$ for attribute $a$ where $N$ is the number of attributes, $B$ is the number of blocks. 
% % Eventually, we sum $N$ attribute predictions' and then $B$ blocks' losses together to obtain the final graph loss $L_{graph}$ similar to Eq (\ref{eq1-3}). 
% Specifically, Eq.~\ref{eq5} demonstrates the process for computing each $Y^{ba}_{graph}$ where $\theta_{ba}$ are learned parameters of classifier $\gamma_{ba}$,  and $\psi$ denotes Sigmoid function.

% \begin{equation} \label{eq5}
% \begin{split}
% Y^{ba}_{graph} & =  \psi (\gamma_{ba}(F^{\prime}_{bi}, \theta_{ba}))
% % L_{graph} & = \sum_{b=1}^{B} \sum_{a=1}^{N} Loss(Y^{ba}_{graph}, Y^{\prime}_a)
% \end{split}
% \end{equation}

\subsection{Multi-Task Prediction and Optimization}
\vspace{-2mm}
\label{loss}
Recalling that our goal is to predict the probability $Y_a$ for each attribute $a$, while our GAL and GCL can generate part-based group features $\f_i^b$ and graph-based group features $\f^{\prime b}_i$. We can predict GAL-based probabilities $Y^{b , GAL}_{a}$ and GCL-based probabilities $Y^{b , GCL}_{a}$ by using $\f_i^b$ and $\f_i^{\prime b}$ for all attributes $a$, \textit{s.t.,} $a \in \mathcal{G_i}$. In addition, we also use the shared feature $\F_B$ extracted from the last block $B$ to obtain a base prediction $Y_{a}^B$ for the attribute $a$.  All prediction processes are denoted as Eq~(\ref{eq3}), where $\gamma$ represents the binary classifier, $\theta$ denotes learned parameters and $\psi$ is the Sigmoid activation function.

\begin{equation} \label{eq3}
\begin{split}
Y^{b,GAL}_{a} & = \psi(\gamma(\f^b_i, \theta^{b}_a)) ~~~~ s.t. ~ a \in \mathcal{G_i} \\
Y^{b,GCL}_{a} & =  \psi (\gamma(\f'^b_i, \theta^{b}_a)) ~~~~ s.t. ~ a \in \mathcal{G_i} \\
Y_{a}^{B} & =  \psi (\gamma(\F_B, \theta^{B}_a)) \\
\end{split}
\end{equation}

The eventual prediction $Y_{a}$ is then calculated by averaging predictions $Y^{b , GCL}_{a}$ from graph-based group feature $\f_i^{\prime b}$ on all blocks and base prediction $Y_{a}^{B}$ from the last shared feature $\F^B$. It is denoted as Eq~(\ref{eq4}), where $B$ is the number of blocks.

\begin{equation} \label{eq4}
\begin{split}
Y_{a} & =  \frac{(Y_a^B + \sum_{b=1}^{B}Y^{b, GCL}_{a})}{1 + B}
\end{split}
\end{equation}

We optimize the entire network by summing all individual prediction lossless $\mathcal{L}$ from each parts. Specifically,
we take $N \times B$ losses from $Y^{b,GAL}_{a}$ 
to optimize GAL, $N \times B$ losses from $Y^{b,GCL}_{a}$ to optimize GCL, $N$ losses from $Y^B_a$ to optimize base prediction and $N$ losses from $Y_{a}$ to optimize final prediction. The total loss is denoted as Eq~(\ref{eq5}), where $N$ is the number of attributes and $\hat{Y_a}$ is the attribute label.

\begin{equation} \label{eq5}
\begin{split}
\mathcal{L}_{total} &= [\sum_{a=1}^{N} \mathcal{L}(Y_{a}, \hat{Y_a}) + \sum_{a=1}^{N} \mathcal{L}(Y_a^B, \hat{Y_a})+ \\
& \sum_{b=1}^{B} \sum_{a=1}^{N}\mathcal{L}(Y^{b,GAL}_{a}, \hat{Y_a}) + \sum_{b=1}^{B} \sum_{a=1}^{N} \mathcal{L} (Y^{b,GCL}_{a}, \hat{Y_a})]
\end{split}
\end{equation}

Importantly, in this work, we aim to enhance both prediction accuracy~\cite{liu2015deep} and balanced accuracy~\cite{huang2019deep}. For prediction optimization, we adopt a binary cross-entropy loss for each $\mathcal{L}$. While for balanced optimization, we employ a weighted binary cross-entropy loss ($\propto$ imbalance level) for each $\mathcal{L}$.

\section{Experiment}
\vspace{-2mm}
\subsection{Experiment Protocols}
\vspace{-1mm}
\textbf{Datasets.} 
We use two public facial attribute datasets CeleBA \cite{liu2015deep} and LFWA \cite{liu2015deep} for evaluation.
CeleBA contains 202,599 facial images with 10K identities. It is split into 162K images for training, 20K images for validation and 20K images for testing. Each image contains 40 attribute labels. 
% There are 10K identities and no identity overlaps in testing, evaluation, and training. Two types of images are provided, which are cropped image and original image. 
It also provides other ground truth data such as landmarks. In our experiments, we only use cropped facial images with attribute labels but not any other supervised data. LFWA is another smaller dataset with 13,233 images and 5,749 identities. It is split into two partitions: the training set with 6,263 images and the testing set with 6,970 images. 
% This dataset is originally selected for identity verification. Hence, all images are well-cropped. 
Each image is annotated with 40 attributes, which are the same as CeleBA. 

\textbf{Evaluation Metrics.} 
We employ two metrics: prediction accuracy and balanced accuracy.
Prediction accuracy is the standard matrix used by previous works \cite{liu2015deep}. This matrix measures the overall performance but somehow ignores the class-imbalanced issue. To demonstrate that our method can also perform well for highly-skewed class distribution, we also adopt the balanced accuracy matrix as in \cite{huang2019deep}. In the next section, we will show mean results of 40 attributes for both metrics on CeleBA and prediction accuracy on LFWA.

\textbf{Implementation.}
\label{p-setting}
To apply our method,
we divide 40 attributes into 8 groups based on their annotations. 
We adopt ResNet18 pre-trained on ImageNet as our backbone and apply GAL and GCL on the last layers of block 3 and block 4. 
For training our MGG-Net, we utilize 3 NVIDIA GTX 1080Ti and set the batch size as 144. The network is trained end-to-end. 
Specifically, for prediction accuracy on CeleBA, we train 9 epochs with learning rate 0.01 and the subsequent 5 epochs with 0.001. For balanced accuracy on CeleBA, we train 12 epochs with learning rate 0.01 and the subsequent 3 epochs with 0.001. For prediction accuracy on LFWA, we train 10 epochs with learning rate 0.01 and the subsequent 5 epochs with 0.001.
The input image is scaled as $224 \times224\times3$ and random flip is used as data argumentation. We also implement a baseline network by only using base prediction results $Y_a^B$ (see details in Section \textcolor{red}{~\ref{loss}}) with above settings.

% \begin{table}[h]
% \resizebox{0.475\textwidth}{!}{
% \begin{tabular}{|c|l|}  
% \hline                      
% \textbf{Groups} & \textbf{Attributes}\\
% \hline
% Mouth & Big lips, Lipstick, Mouth slightly open, Smiling, Goatee,\\
% & 5 o'clock shadow, Mustache, Double chin, Beard\\
% \hline
% Eyes & Arched eyebrows, Bushy eyebrows, Eye glasses, Narrow eyes \\
% \hline
% Whole Face & Attractive, Blurry, Makeup, Oval, Young, Male, Chubby, Pale, \\
% \hline
% Hairline & Bald, Bangs, Receding Hairline, Hat \\
% \hline
% Around Head & Black hair, Blond hair, Brown hair, \\
% & Gray hair, Straight hair, Wavy hair \\
% \hline
% Middle Face & High cheekbones, Rosy cheeks,  Eye Bags, Sideburns, Earrings \\
% \hline
% Nose & Big nose, Pointy nose \\
% \hline
% Neck &  Necklace, Necktie \\
% \hline
% \end{tabular}}
% \caption{We divide 40 attributes into 8 part-based groups.}
% \label{tab:group}
% \end{table}

% \vspace{-5mm}

\subsection{Experiment Results}
\vspace{-2mm}
\textbf{Prediction Accuracy.}
Table~\ref{Tab:celeba} reports the mean prediction accuracy results of 40 attributes on CeleBA and LFWA. In fact, most reported methods utilized extra data besides attribute labels. For example, PANDA~\cite{zhang2014panda} and FaceTracer~\cite{kumar2008facetracer} adopted landmarks.
Walk and Learn~\cite{wang2016walk} obtained 5 million online sources. Off-the-Shell~\cite{zhong2016face} used another large face recognition dataset. Kalayeh~\etal~\cite{kalayeh2017improving} and He~\etal~\cite{he2018harnessing} trained their segmentation and generation networks on a large face parsing domain. Differently, our approach does not rely on any auxiliary ground-truth data and our network is trained end-to-end. 

\begin{table}[h!]
  \begin{center}
   \resizebox{0.48\textwidth}{!}{
    \begin{tabular}{c|c|c} 
      \textbf{Method} & \textbf{CeleBA (\%)} & \textbf{LFWA (\%)}\\ 
      \hline
      \hline
      (Kumar \etal 2009) FaceTracer \cite{kumar2008facetracer} & 81.12  &  73.93\\ \hline
      (Zhang \etal 2014) PANDA \cite{zhang2014panda} & 85.00 & 81.00\\ \hline
      (Liu \etal 2015) ANet+LNet \cite{liu2015deep} & 87.30 & 84.00 \\ \hline
      (Zhong \etal 2016) Off-the-Shell \cite{zhong2016face} & 86.60 & 85.90 \\ \hline
      (Wang \etal 2016) Walk and Learn \cite{wang2016walk} & 88.00 & -\\ \hline
      (Rudd \etal 2016) Separate \cite{rudd2016moon} & 90.22 & -\\ \hline
      (Rudd \etal 2016) MOON \cite{rudd2016moon} & 90.94 & - \\ \hline
      (Hand \etal 2017) MCNN-AUX \cite{hand2017attributes} & 91.26 & 86.30 \\ \hline
      (Lu \etal 2017) SOMP joint branch-32 \cite{lu2017fully} & 90.40 & - \\ \hline
      (Lu \etal 2017) SOMP joint branch-64 \cite{lu2017fully} & 91.02 & - \\ \hline
      (Kalayeh \etal 2017) Avg Pooling \cite{kalayeh2017improving} & 90.86 & 85.27\\ \hline
      (Kalayeh \etal 2017) SSG \cite{kalayeh2017improving}  & 91.62 & 86.13 \\ \hline
      (Kalayeh \etal 2017) SSP \cite{kalayeh2017improving} & 91.67   & 86.80 \\ \hline
      (Kalayeh \etal 2017) SSG + SSP \cite{kalayeh2017improving} & 91.80 & 87.13 \\ \hline
      (Huang \etal 2018) GNAS (Thin) \cite{huang2018gnas} & 91.30 & 86.16 \\ \hline
      (Huang \etal 2018) GNAS (Wide) \cite{huang2018gnas} & 91.63 & 86.37 \\  \hline
      (He \etal 2018) Original \cite{he2018harnessing} & 91.50   & 84.79 \\ \hline
      (He \etal 2018) Abstract + Original \cite{he2018harnessing} & 91.81 & 85.28\\ \hline
      (Zhao \etal 2019) CSN w/o share \cite{zhao2019recognizing} & 91.70 & - \\ \hline
      (Zhao \etal 2019) CSN \cite{zhao2019recognizing} & 91.80  & -\\ 
      \hline
      \hline
      Ours: Baseline & 91.23 & 85.63 \\
      Ours: MGG-Net & \textbf{92.00} & \textbf{87.20}
    \end{tabular}
   }
  \end{center}
  \caption{\label{Tab:celeba}Mean prediction accuracy (in \%) of 40 attributes on CeleBA and LFWA compared to state-of-the-art methods.}
\end{table}

Our baseline and MGG-Net contain 11M and 12.8M parameters. Thanks to the group structure, it totally costs 1.8M parameters for GAL and GCL only. With 16\% additional parameters, our approach reduces 9.6\% and 10.92\% relative classification errors on CeleBA and LFWA respectively. Compared to ours, MOON~\cite{rudd2016moon} had 119.73M parameters and ANet+LNet~\cite{liu2015deep} costed 128M parameters. He \etal~\cite{he2018harnessing} took ResNet50 which had
at least 50M parameters. SSG+SSP~\cite{kalayeh2017improving} and MCNN-Aux~\cite{hand2017attributes} had around 24M and 16M parameters. Though our parameters cost is slightly larger than SOMP~\cite{lu2017fully} (10.53M), it is still fair to claim that MGG-Net achieves the highest results using a relatively light-weight setting.

In terms of improvement on CeleBA, most methods enhanced from baselines by 0.3\% - 0.9\% for mean prediction accuracy. Specifically, MOON \cite{rudd2016moon} outperformed its separated method by 0.72\%. SSG and SSP \cite{kalayeh2017improving} enhanced 0.76\% and 0.81\% respectively from only using average pooling. Abstract method \cite{he2018harnessing} also contributed a 0.31\% points gain from only using original images. While our proposed MGG-Net achieves 0.77\% enhancement from our baseline network. 
For LFWA, it is observed that MGG-Net makes 1.57\% points improvement from the baseline network and has outstanding performance against all other methods. Compared to the most recent state-of-the-art approach that used abstract images \cite{he2018harnessing}, we even improve 1.92\%.
Moreover,  our MGG-Net is also significantly superior over other MTL-based approaches on both CeleBA and LFWA, which are hand-designed grouping method MCNN-Aux \cite{hand2017attributes}, adaptive correlation learning method SOMP \cite{lu2017fully} and greedy search method GNAS \cite{huang2018gnas}.

\textbf{Balanced Accuracy.}
We also demonstrate the effectiveness of our approach in the balanced setting. 
We compare our approach with four recent deep imbalance learning methods LMLE~\cite{huang2019deep}, CRL \cite{dong2018imbalanced}, CLMLE \cite{huang2019deep} and DCL \cite{wang2019dynamic} on CeleBA dataset. 
To our best knowledge, there are no existing methods evaluated the balanced task on LFWA dataset.

\begin{table}[h!]
  \begin{center}
    \label{tab:celebablance}
    \resizebox{0.4\textwidth}{!}{
    \begin{tabular}{c|c} % <-- Alignments: 1st column left, 2nd middle and 3rd right, with vertical lines in between
      \textbf{Method} & \textbf{Balanced Accuracy (\%)} \\ 
      \hline
      \hline
    %   (Zhang et al 2014) PANDA \cite{zhang2014panda} & 76.95 \\ \hline
    %   (Liu et al 2015) LNets+ANets \cite{liu2015deep} & 79.58 \\ \hline
    %   (Rudd et al 2016) MOON \cite{rudd2016moon} & 78.59 \\ \hline
      (Huang \etal 2016) LMLE \cite{huang2019deep} & 83.83 \\ \hline
      (Dong \etal 2017) CRL \cite{dong2018imbalanced} & 86.60 \\ \hline
      (Huang \etal 2018) CLMLE \cite{huang2019deep} & 88.78 \\ \hline
      (Wang \etal 2019) DCL \cite{wang2019dynamic} & 89.05 \\
      \hline
      \hline
      Ours: Baseline: & 88.15 \\
      Ours: MGG-Net: & \textbf{89.19}
      
    \end{tabular}}
  \end{center}
  \caption{\label{Tab:C}Mean balanced accuracy (in \%)  of 40 attributes on CeleBA dataset compared to other state-of-the-art methods.}
\end{table}

As shown in Table~\ref{Tab:C},  our MGG-Net achieves the highest mean balanced accuracy 89.19\% and enhances 1.04\% from the baseline. 
We highlight two facts.
First, these four methods all had high complexity. For instance, the most recent state-of-the-art method DCL~\cite{wang2019dynamic} re-sampled its scheduler in every iteration during the training and took 300 epochs with batch size 512 for converging. While ours adopts a group-structure, which significantly reduces the complexity and only takes 15 epochs with batch size 144 for training.
Second, all existing methods were only proposed for either the balanced or prediction accuracy (\textit{e.g.,} DCL was only optimized for enhancing balanced accuracy but not for prediction accuracy). While our approach is able to achieve excellent results on both evaluation metrics by simply switching the training criterion.

\subsection{Ablation Study}
\vspace{-2mm}
\textbf{Visualization of Group Attentions.} We visualize the group attention masks to study the effectiveness of GAL. In our implementation, we extract eight $28 \times 28 \times 1$ masks from block 3 and eight $14 \times 14 \times 1$ masks from block 4 of ResNet18. 

\begin{figure}[h]
\begin{center}
   \includegraphics[width=0.9\linewidth]{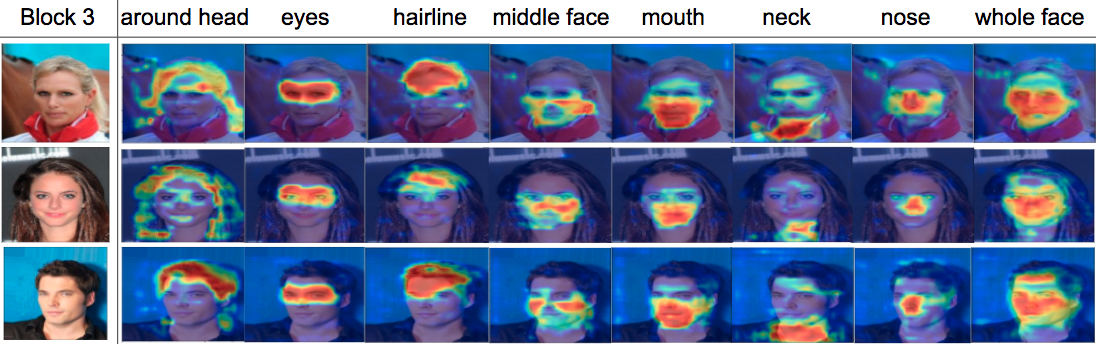}
\end{center}
\begin{center}
   \includegraphics[width=0.9\linewidth]{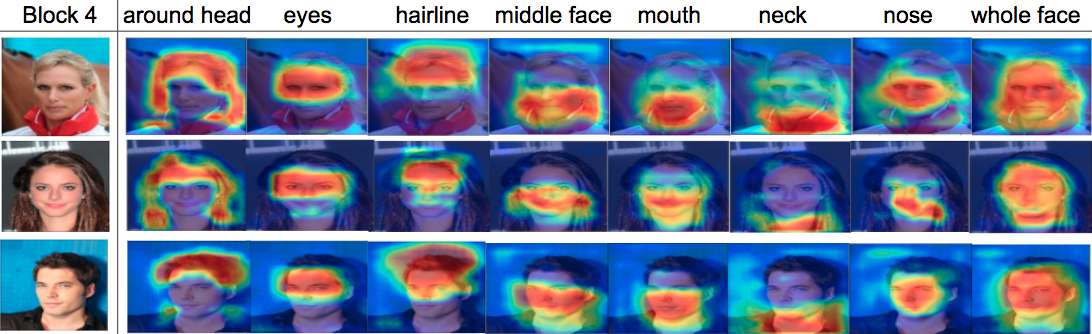}
\end{center}
   \caption{Visualization of output group attention masks on block 3 ($28 \times 28 \times 1$) and block 4 ($14 \times 14 \times 1$) of ResNet18.}
\label{fig:groupvis}
\end{figure}

Fig.~\ref{fig:groupvis} shows output masks on each block. It is observed that attentions on block 3 can better outline the finer and smaller regions, which correspond to eyes, mouth, nose, and hairline. While masks on block 4 perform better on the coarser and larger regions which are the whole face, around head, neck and middle face. It indicates that coarse-to-fine multi-scale design can benefit different types of facial parts. 

% We visualize the generated spatial attentions on each block. 

% We also observe that both block 3 and 4 can detect the facial parts we design for groups. 

\textbf{Effectiveness of GAL and GCL.}
To demonstrate the effectiveness of GAL and GCL, we further implement two networks:
the first one learns all attribute attentions independently (Independent) and the second one only applies group attentions (w/o GCL).
We use the same design choice and training scheme as MGG-Net (see \textit{Implementation}) for evaluating them on LFWA. Results are shown in Table~\ref{tab:ab1}.

\begin{table}[h!]
  \begin{center}
    \resizebox{0.45\textwidth}{!}{
    \begin{tabular}{c|c|c|c} % <-- Alignments: 1st column left, 2nd middle and 3rd right, with vertical lines in between
      \textbf{Layer} & \textbf{Independent (\%)} & \textbf{w/o GCL(\%)} & \textbf{MGG-Net(\%)} \\ 
      \hline
      \hline
      Block 3 & 81.04 & 82.40 & \textbf{86.40}\\ \hline
      Block 4 & 85.27 & 86.42 & \textbf{86.81}\\ \hline
      Block 3+4 & 86.17 & 86.77 & \textbf{87.20}\\ \hline
    \end{tabular}}
  \end{center}
  \caption{\label{tab:ab1} Mean prediction accuracy (in \%) on LFWA dataset by using independent attentions (Independent), only using GAL (w/o GCL), and using both GAL and GCL (MGG-Net).}
\end{table}

It shows that GAL enhances the performance from using independent attentions by more than 1\% on both block 3 and block 4. There are 40 attributes but only 8 groups, so that using independent attentions costs extra 4 times parameters than using group attentions.
Hence, it indicates that GAL not only reduces parameters, but also improves performance.
We also see that results on both block 3 and block 4 are enhanced by further collaborating with GCL. Moreover, GCL helps to improve overall performance (block 3 + 4 improves 0.43\%).

\begin{table}[h!]
  \begin{center}
    \resizebox{0.42\textwidth}{!}{
    \begin{tabular}{c|c|c|c} % <-- Alignments: 1st column left, 2nd middle and 3rd right, with vertical lines in between
      \textbf{Backbone} & \textbf{Baseline (\%)} & \textbf{w/o GCL (\%)} & \textbf{MGG-Net (\%)} \\
    %   \textbf{bone} & \textbf{(\%)} & \textbf{(\%)} & \textbf{Graph (\%)} \\
      \hline
      \hline
      Resnet18 & 85.63 & 86.77 & \textbf{87.20}\\ \hline
      Resnet34 & 85.65 & 86.42 & \textbf{86.81} \\ \hline
      VGG16 & 84.57 & 84.06 & \textbf{85.22} \\ \hline
      AlexNet & 84.30 & 84.90 & \textbf{85.59} \\ \hline
    \end{tabular}}
  \end{center}
  \caption{\label{tab:ab3} Mean prediction accuracy (in \%) on LFWA with different backbones for our baseline, the network only using GAL (w/o GCL), and MGG-Net using both GAL and GCL.}
\end{table}

We also show the effectiveness of our method by switching different backbones for our baseline, the network only using GAL and our MGG-Net using both GCL and GAL. The results in Table~\ref{tab:ab3} indicate that our GCL and GAL consistently boost the accuracy with different backbones. While simply increasing parameters by changing backbones or only using GAL does not necessarily improve the performance.

% \textbf{Analysis of Design Choice}. Lastly, we show results with different designs in Table~\ref{Tab:gcnc}. It shows that introducing more blocks to apply our method can further improve the performance. Yet, the 3-blocks design will also bring more parameters to our model. Therefore, we only utilize 2 blocks to achieve a fair trade-off between accuracy and complexity.

% \begin{table}[h!]
%   \begin{center}
%     \resizebox{0.35\textwidth}{!}{
%     \begin{tabular}{c|c|c} % <-- Alignments: 1st column left, 2nd middle and 3rd right, with vertical lines in between
%       \textbf{Design} & \textbf{Accuracy (\%)} & \textbf{\#Parameter (M)} \\
%     %   \textbf{bone} & \textbf{(\%)} & \textbf{(\%)} & \textbf{Graph (\%)} \\
%       \hline
%       \hline
%       Baseline  & 85.63 & 11 \\ \hline
%       Block 3 & 86.67 & 11.6  \\ \hline
%       Block 3+4 & 87.20 & 12.8  \\ \hline
%       Block 3+4+2 & 87.27 & 13.2  \\ \hline
%       Block 3+4+5 & 87.31 & 17.5  \\ \hline
%     \end{tabular}}
%   \end{center}
%   \caption{\label{Tab:gcnc} Mean prediction accuracy (in \%) on LFWA and parameter numbers (in M) of MGG-Net with different designs.} 
% \end{table}

\vspace{-3mm}
\section{Conclusion}
\vspace{-2mm}
In this work, we have proposed a novel method to enhance facial attribute recognition
by exploiting and leveraging both spatial and non-spatial correlations between attributes.
Our MGG-Net achieves new state-of-the-art results in the evaluation. Moreover, the ablation study shows that GAL is efficient to obtain good part-based group features by intra-group spatial similarity, while GCL can further enhance group features by inter-group non-spatial affinity.
Overall, by better handling correlations between multiple facial attributes, our method is sufficient to improve the holistic recognition performance.

\bibliographystyle{IEEEbib}
\bibliography{icme2020}

\end{document}

\sloppy

% Example definitions.
% --------------------
\def\x{{\mathbf x}}
\def\L{{\cal L}}

% Title.
% ------
\title{Improving Facial Attributes Recognition by Group and Graph Learning \\
      (Supplementary Material)}
%
% Single address.
% ---------------
\name{\vspace{-2mm}Zhenghao Chen$^\dag$, Shuhang Gu$^\dag$, Feng Zhu$^\ddag$, Jing Xu$^*$ and Rui Zhao$^\ddag$}

%Address and e-mail should NOT be added in the submission paper. They should be present only in the camera ready paper. 
\vspace{-2mm} 
\address{$^\dag$\textit{The University of Sydney},
          $^\ddag$\textit{Sensetime Group Limited and} 
          $^*$\textit{Applied Research Center, Tencent PCG} \\
         \small{zhenghao.chen@sydney.edu.au, shuhanggu@gmail.com, \{zhufeng, zhaorui\}@sensetime.com and
          eudoraxu@tencent.com}}

\maketitle

\begin{abstract}
This document has the following sections:
(A) Architecture of Group Attention Modules in GAL. 
(B) Details of evaluation metrics for prediction and balanced accuracy. 
(C) Details of individual objective losses for prediction and balanced optimization. 
(D) The design choice of part-based groups for CeleBA and LFWA datasets.
(E) Individual accuracy results for 40 attributes on CeleBA and LFWA datasets. 
(F) Attention masks visualization of independent attention learning and GAL. 
(G) Visualization of group affinity matrix learned from GCL.
(H) Settings of different backbone networks.
(I) Analysis of design choice for our MGG-Net.
\end{abstract}

\section{Group Attention Module}
\begin{figure}[!h]
\begin{center}
  \includegraphics[width=0.8\linewidth]{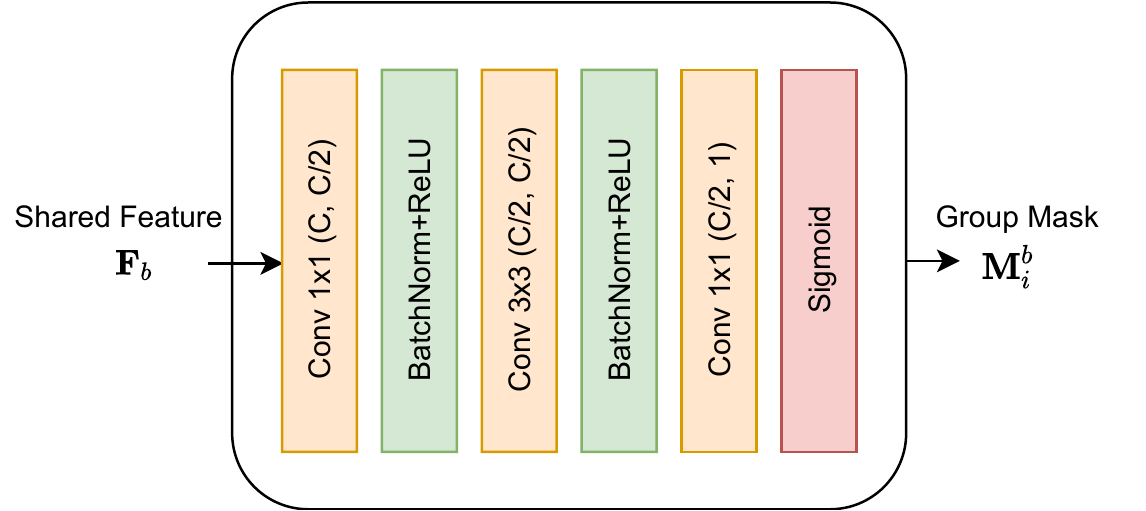}
\end{center}
  \caption{Group Attention Modules for GAL. Conv $n\times n$ ($C_1$, $C_2$) represents $n \times n$ convolution kernel with input channel $C_1$ and output channel $C_2$. 
  The stride is set to 1 for all convolution operations and $C$ is the input channel number.}
\label{fig:groupattention}
\end{figure}

\thispagestyle{FirstPage}
The architecture of Group Attention Modules is shown in Fig.~\ref{fig:groupattention}.
It takes the shared feature $\F_b$ extracted from the block $b$ as the input and then produce group mask $\M_i^b$ as the output for group $\mathbb{G}_i$.
In our implementation, we adopt the last layers of block 3 and block 4 from ResNet18 to apply GAL.
For block 3, the input channel number $C$ is 128. While for block 4, the input channel number $C$ is 512. We set the stride to 1 for all convolution operations in the Group Attention Module.

\section{Evaluation Matrix}
\vspace{-2mm}
In our experiments, we employ two evaluation metrics that are mean prediction accuracy and mean balanced accuracy.

For prediction accuracy, we follow the setting as in \cite{liu2015deep} and compute the value for attribute $a$ by using the number of correct predictions $T_a$ to divide the number of samples $T$.

For balanced accuracy, we follow the setting as in \cite{huang2019deep}. We first compute the accuracy of positive samples ($P'_a / P_a$) and the accuracy of negative samples ($I'_a / I_a$) for attribute $a$. Then, we further obtain the mean of these two values. $P'_a$ and $I'_a$ are the numbers of correct predictions for positive samples and negative samples, while $P_a$ and $I_a$ are the numbers of positive samples and negative samples. 

Lastly, we take the means of prediction accuracy and balanced accuracy respectively over $N$ attributes as our final experimental results demonstrated in the main paper.
The process is denoted as Eq~(\ref{eq1}). 

\begin{equation} \label{eq1}
\begin{split}
Mean\ Prediction\ & = \frac{1}{N} \sum^{N}_{a=1} (T_a / T)  \\
Mean\ Balanced\ &  = \frac{1}{N} \sum^{N}_{a=1} (P'_a / P_a + I'_a / I_a) / 2
\end{split}
\end{equation}

\\

\section{Individual Objective Loss}
In the main paper, we add up $2N \times (B+1)$ individual losses $\mathcal{L}$($Y'_a$, $\hat{Y}_a$) for our total loss, where $Y'_a$ is the prediction results for attribute $a$, $\hat{Y}_a$ is the attribute label, $N$ is the number of attributes and $B$ is the number of blocks. 
Specifically, we have four types of prediction results, which are $Y^{b,GCL}_a$, $Y^{b,GAL}_a$, $Y^{B}_a$ and $Y_a$.
To optimize our network for both prediction accuracy and balanced accuracy, we employ two different objective functions for the individual loss $\mathcal{L}(Y'_a, \hat{Y}_a)$.

For optimizing the prediction accuracy, we directly use a binary cross-entropy loss to fairly count the loss for each sample, which is denoted as Eq~(\ref{eq2}). 

\begin{equation} \label{eq2}
\begin{split}
\mathcal{L}(Y'_a, \hat{Y}_a) = -log(Y'_{a})\hat{Y}_a - log(1 - Y'_{a})(1- \hat{Y}_a)   
\end{split}
\end{equation}

\\

For imbalanced attributes, there are only a few samples in the minority class. It will be hard to learn the minority class if we directly employ binary cross-entropy loss, which forces every sample to contribute equally in the back-propagation process. Therefore, we employ a weighted binary cross-entropy loss for optimizing the balanced accuracy, which is denoted as Eq~(\ref{eq3}), where $S_a$ is the number of samples with the predicting class in one batch and $S$ represents the batch size. 
In every training iteration, this loss will put more effort into the minority class at batch-wise level. 

\begin{equation} \label{eq3}
\begin{split}
\mathcal{L}(Y'_a, \hat{Y}_a) &= - log(Y'_{a})\hat{Y}_a \frac{S-S_a}{S} \\
& - log(1 - Y'_{a})(1- \hat{Y}_a) \frac{S_a}{S}   
\end{split}
\end{equation}

\\

\section{Part-based Group}
In our experiment, we divide 40 attributes from CeleBA and LFWA datasets into 8 part-based groups based on their annotations. We adopt the identical design choice for experiments on both CeleBA and LFWA datasets.\\

\begin{table}[h]
\centering

\resizebox{0.48\textwidth}{!}{
\begin{tabular}{c|l}  
\textbf{Groups} & \textbf{Attributes}\\
\hline \hline
& Big lips, Lipstick, Mouth slightly open, \\
Mouth & Smiling, Goatee, 5 o'clock shadow, Mustache, \\
& Double chin, Beard \\
\hline
Eyes & Arched eyebrows, Bushy eyebrows, \\
& Eye glasses, Narrow eyes \\
\hline
Whole face & Attractive, Blurry, Makeup, Oval, \\
& Young, Male, Chubby, Pale \\
\hline
Hairline & Bald, Bangs, Receding hairline, Hat \\
\hline
Around head & Black hair, Blond hair, Brown hair, \\
& Gray hair, Straight hair, Wavy hair \\
\hline
Middle face & High cheekbones, Rosy cheeks,  Eye bags, \\
& Sideburns, Earrings \\
\hline
Nose & Big nose, Pointy nose \\
\hline
Neck &  Necklace, Necktie
\end{tabular}}
\label{tab:group}
\caption{We divide 40 attributes into 8 part-based groups. This design choice is used for both CeleBA and LFWA.}
\end{table}
 \\

\section{Individual Performance over 40 attributes}
We compute the prediction accuracy and balanced accuracy for all 40 attributes on CeleBA and LFWA datasets. However, we do not show the performance for all 40 attributes individually in the paper due to the limit of pages. In this section, we will provide the details of individual experimental results for all attributes. Table~\ref{tab:index} shows the attribute index for both CeleBA dataset and LFWA dataset.  \\

\begin{table}[!h]
\centering

\resizebox{0.45\textwidth}{!}{
\begin{tabular}{ll|ll}
1  & 5 o'Clock shadow & 21 & Male                \\
2  & Arched eyebrows  & 22 & Mouth slightly open \\
3  & Attractive       & 23 & Mustache            \\
4  & Bags under eyes  & 24 & Narrow eyes         \\
5  & Bald             & 25 & No beard            \\
6  & Bangs            & 26 & Oval face           \\
7  & Big lips         & 27 & Pale skin           \\
8  & Big nose         & 28 & Pointy nose         \\
9  & Black hair       & 29 & Receding hairline   \\
10 & Blond hair       & 30 & Rosy cheeks         \\
11 & Blurry           & 31 & Sideburns           \\
12 & Brown hair       & 32 & Smiling             \\
13 & Bushy eyebrows   & 33 & Straight hair       \\
14 & Chubby           & 34 & Wavy hair           \\
15 & Double chin      & 35 & Wearing earrings    \\
16 & Eye glasses      & 36 & Wearing hat         \\
17 & Goatee           & 37 & Wearing lipstick    \\
18 & Gray hair        & 38 & Wearing necklace    \\
19 & Heavy makeup     & 39 & Wearing necktie     \\
20 & High cheekbones  & 40 & Young              \\
\end{tabular}}
\caption{\label{tab:index} Attribute index for CeleBA and LFWA datasets.}
\end{table}

\\

Table~\ref{tab:celeba-a} reports the prediction accuracy on CeleBA, Table~\ref{tab:lfwa-a} reports the prediction accuracy on LFWA and Table~\ref{tab:celeba-b} demonstrates the balanced accuracy on CeleBA.
For prediction accuracy, we list ANet+LNet \cite{liu2015deep} and MCNN-Aux \cite{hand2017attributes} to compare with our approach. For balanced accuracy, we list CRL \cite{dong2018imbalanced} and CLMLE \cite{huang2019deep} for the evaluation.

In Table~\ref{tab:celeba-a}, it is observed that our method achieves the best results for 29 attributes on CeleBA dataset. 
Table~\ref{tab:lfwa-a} shows that our method achieves the best results for 24 attributes on LFWA dataset.
In terms of Table~\ref{tab:celeba-b}, the results are ordered by the imbalance level of attributes from CeleBA \cite{huang2019deep}. That is, attribute 3 (``attractive") has the most balanced distribution of samples, while attribute 5 (``bald") has the most imbalanced distribution of samples. It is observed that our method MGG-Net achieves the best balanced accuracy results for 20 attributes among all approaches. 
Also, we can see that our method MGG-Net enhances performance from the baseline network for almost all attributes on evaluation results.

Overall, the experimental results show that our method not only achieves the best performance in terms of mean prediction accuracy and mean balanced accuracy comparing to other state-of-the-art methods as shown in the main paper, but also obtain a significantly superior performance regarding the individual attribute prediction accuracy and balanced accuracy over other methods.
\begin{table*}[!h]
\centering
\resizebox{\textwidth}{!}{
\begin{tabular}{c|cccccccccccccccccccc}
Attribute Index & 1           & 2           & 3           & 4           & 5           & 6           & 7           & 8           & 9           & 10          & 11          & 12          & 13          & 14          & 15          & 16           & 17           & 18          & 19          & 20          \\ \hline
ANet+LNet \cite{liu2015deep} & 91          & 79          & 81          & 79          & 98          & 95          & 68          & 78          & 88          & 95          & 84          & 80          & 90          & 91          & 92          & 82           & 99           & 95          & 97          & 99          \\
MCNN-Aux \cite{hand2017attributes}  & 95          & 83          & 83          & 85          & 99          & 96          & 71          & 85          & 90          & 96          & 96          & 89          & 93          & 96          & 96          & 90           & \textbf{100} & 97          & \textbf{98} & \textbf{99} \\
Ours: Baseline   & 95          & 84          & 82          & 84          & 99          & 96          & 72          & 83          & 89          & 96          & 96          & 89          & 92          & 96          & 96          & 100          & 97           & 98          & 92          & 87          \\
Ours: MGG-Net   & \textbf{95} & \textbf{85} & \textbf{84} & \textbf{86} & \textbf{99} & \textbf{96} & \textbf{73} & \textbf{85} & \textbf{91} & \textbf{96} & \textbf{96} & \textbf{90} & \textbf{93} & \textbf{96} & \textbf{97} & \textbf{100} & 98           & \textbf{98} & 92          & 88        \\  \\ 
\hline  \hline \\
Attribute Index & 21          & 22          & 23          & 24          & 25          & 26          & 27          & 28          & 29          & 30          & 31          & 32          & 33          & 34          & 35          & 36           & 37           & 38          & 39          & 40          \\ \hline
ANet+LNet \cite{liu2015deep} & 90          & 88          & 93          & 98          & 92          & 95          & 81          & 71          & 93          & 95          & 66          & 91          & 72          & 89          & 90          & 96           & 92           & 73          & 80          & 87          \\
MCNN-Aux \cite{hand2017attributes} & 92          & 88          & 94          & \textbf{98} & 94          & \textbf{97} & 87          & \textbf{87} & \textbf{97} & \textbf{96} & 76          & \textbf{97} & 77          & \textbf{94} & \textbf{95} & 98           & 93           & 84          & 84          & 88          \\
Ours: Baseline    & 98          & 94          & 97          & 87          & 96          & 75          & 97          & 76          & 94          & 95          & 98          & 93          & 84          & 85          & 91          & 99           & 94           & 88          & 97          & 88          \\
Ours: MGG-Net   & \textbf{99} & \textbf{94} & \textbf{97} & 88          & \textbf{97} & 77          & \textbf{97} & 78          & 94          & 95          & \textbf{98} & 93          & \textbf{85} & 86          & 91          & \textbf{99}  & \textbf{95}  & \textbf{88} & \textbf{97} & \textbf{89}
\end{tabular}}
\caption{Prediction accuracy results (in \%) of 40 attributes on CeleBA.}
\label{tab:celeba-a}
\end{table*}

\begin{table*}[!h]
\centering
\resizebox{\textwidth}{!}{
\begin{tabular}{c|cccccccccccccccccccc}
Attribute Index & 1           & 2           & 3           & 4           & 5           & 6           & 7           & 8           & 9           & 10          & 11          & 12          & 13          & 14          & 15          & 16          & 17          & 18          & 19          & 20          \\ \hline
ANet+LNet \cite{liu2015deep} & \textbf{84} & 82          & \textbf{83} & 83          & 88          & 88          & 75          & 81          & 90          & 97          & 74          & 77          & 82          & 73          & 78          & 94          & \textbf{95} & 78          & 84          & 88          \\
MCNN-Aux \cite{hand2017attributes} & 77          & 82          & 80          & 83          & 92          & 90          & 79          & 85          & \textbf{93} & 97          & 85          & 81          & 85          & 77          & 82          & \textbf{95} & 91          & 83          & 89          & \textbf{90} \\
Ours: Baseline    & 77          & 81          & 79          & 81          & 92          & 90          & 78          & 83          & 92          & 97          & 86          & 80          & 83          & 75          & 81          & 91          & 83          & 88          & 96          & 88          \\
Ours: MGG-Net   & 79          & \textbf{83} & 81          & \textbf{85} & \textbf{93} & \textbf{92} & \textbf{80} & \textbf{85} & 92          & \textbf{98} & \textbf{88} & \textbf{83} & \textbf{86} & \textbf{78} & \textbf{82} & 92          & 85          & \textbf{89} & \textbf{96} & 89          \\
\\ \hline \hline \\ 
Attribute Index & 21          & 22          & 23          & 24          & 25          & 26          & 27          & 28          & 29          & 30          & 31          & 32          & 33          & 34          & 35          & 36          & 37          & 38          & 39          & 40          \\ \hline
ANet+LNet \cite{liu2015deep} & 95          & 88          & 95          & 94          & 82          & 92          & 81          & 88          & 79          & 79          & 74          & 84          & 80          & 85          & 78          & 77          & 91          & 76          & 76          & 86          \\
MCNN-Aux \cite{hand2017attributes} & \textbf{96} & \textbf{88} & \textbf{95} & \textbf{94} & \textbf{84} & \textbf{93} & 83          & \textbf{90} & 81          & 82          & 77          & \textbf{93} & \textbf{84} & \textbf{86} & 88          & 83          & 92          & 79          & 82          & 86          \\
Ours: Baseline    & 93          & 82          & 93          & 81          & 82          & 76          & 90          & 83          & 85          & 86          & 82          & 91          & 81          & 80          & 95          & 90          & 95          & 90          & 82          & 86          \\
Ours: MGG-Net   & 94          & 84          & 94          & 84          & 83          & 79          & \textbf{91} & 84          & \textbf{86} & \textbf{88} & \textbf{84} & 92          & 83          & 82          & \textbf{95} & \textbf{91} & \textbf{95} & \textbf{90} & \textbf{84} & \textbf{87}
\end{tabular}}
\caption{\label{tab:lfwa-a} Prediction accuracy results (in \%) of 40 attributes on LFWA.}
\end{table*}

\begin{table*}[!h]
\centering
\resizebox{\textwidth}{!}{
\begin{tabular}{c|cccccccccccccccccccc}
Attribute Index & 3           & 22          & 32          & 37          & 20          & 21          & 19          & 34          & 26          & 28          & 2           & 9           & 7           & 8           & 40          & 33          & 12          & 4           & 35          & 25          \\ \hline
CRL \cite{dong2018imbalanced}            & 83          & 95          & 93          & 94          & 89          & 96          & 84          & 79          & 66          & 73          & 80          & 90          & 68          & 80          & 84          & 73          & 86          & 80          & 83          & 94          \\
CLMLE \cite{huang2019deep}          & 90          & \textbf{97} & \textbf{99} & \textbf{98} & 94          & \textbf{99} & \textbf{98} & 87          & 72          & 78          & 86          & \textbf{95} & 66          & 85          & 90          & 80          & 89          & 82          & 86          & \textbf{98} \\
Baseline          & 93          & 94          & 89          & 91          & 95          & 95          & 88          & 91          & 91          & 94          & 94          & 69          & 92          & 84          & 97          & 96          & 89          & 94          & 98          & 76          \\
MGG-Net         & \textbf{94} & 94          & 89          & 93          & \textbf{96} & 95          & 89          & \textbf{92} & \textbf{93} & \textbf{94} & \textbf{94} & 72          & \textbf{92} & \textbf{85} & \textbf{98} & \textbf{96} & \textbf{90} & \textbf{95} & \textbf{98} & 79          \\
                &             &             &             &             &             &             &             &             &             &             &             &             &             &             &             &             &             &             &             &             \\ \hline \hline \\
Attribute Index & 6           & 10          & 13          & 38          & 24          & 1           & 29          & 39          & 16          & 30          & 17          & 14          & 31          & 11          & 36          & 15          & 27          & 18          & 23          & 5           \\ \hline
CRL \cite{dong2018imbalanced}            & 95          & 95          & 84          & 74          & 72          & 90          & 87          & 88          & 96          & 88          & 96          & 87          & 92          & \textbf{85} & 98          & \textbf{89} & 92          & 95          & \textbf{94} & \textbf{97} \\
CLMLE \cite{huang2019deep}          & 99 & \textbf{99} & \textbf{88} & 69          & 71          & \textbf{91} & 82          & 96          & \textbf{99} & 86          & \textbf{98} & \textbf{85} & 94          & 72          & \textbf{99} & 87          & \textbf{94} & \textbf{96} & 82          & 95          \\
Baseline          & 99          & 72          & 68          & 94          & 78          & 82          & 99          & 95          & 80          & 91          & 85          & 79          & 96          & 83          & 89          & 85          & 87          & 82          & 87          & 87          \\
MGG-Net         & \textbf{99}          & 74          & 70          & \textbf{96} & \textbf{79} & 83          & \textbf{99} & \textbf{96} & 81          & \textbf{91} & 86          & 80          & \textbf{96} & 84          & 90          & 86          & 89          & 83          & 88          & 88         

\end{tabular}}
\caption{\label{tab:celeba-b} Balanced accuracy results (in \%) of 40 attributes on CeleBA.}
\end{table*}

\section{Individual and Group Attention}

\begin{figure}[!h]
\begin{center}
   \includegraphics[width=0.9\linewidth]{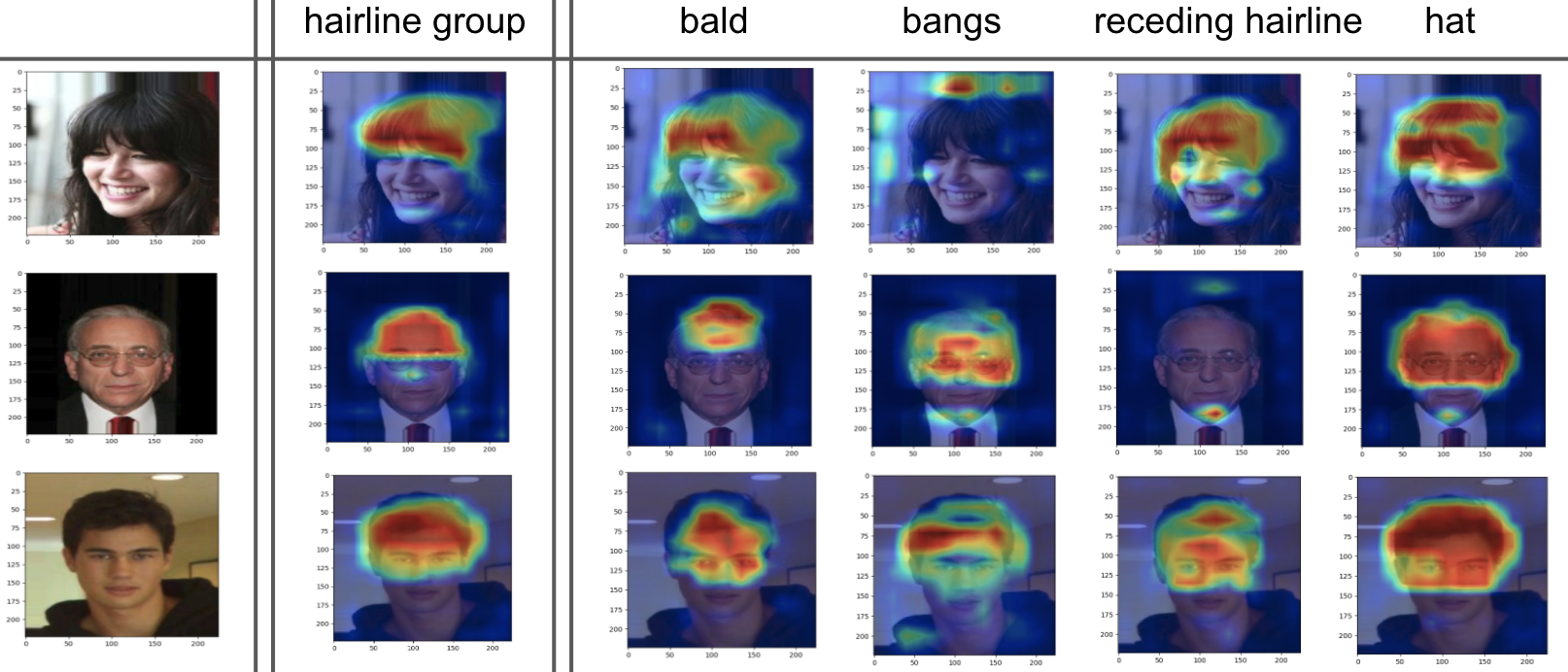}
\end{center}
   \caption{Visualization of the group and individual attentions generated from block 4 of ResNet18 for the ``hairline" group. Failure cases (attributes) show the difficulty of directly capturing the attention for hard-case attributes (\textit{e.g.,} ``bangs" and ``receding hairline") independently. However, our proposed GAL can mitigate this issue by leveraging the spatial correlation between attributes. We see that group attention masks can well-cover the hairline region.}
\label{fig:groupvis2}
\end{figure}
\vspace{-2mm}
In the ablation study, we implement a network that learns part-based attentions independently for each attribute without using GAL and we compare this network to our GAL. 
The results in the paper indicate that our GAL achieves better overall performance by leveraging intra-group spatial similarity between attributes.
To further demonstrate the effectiveness of GAL, we visualize attention masks for both individual and group attentions generated from block 4 of ResNet18. 
We show the visualization in Fig.~\ref{fig:groupvis2}. 

Specifically, we take the ``hairline" group (with attributes ``bald", ``bangs", ``receding hairline" and ``hat") as an example. This visualization reveals that it fails to obtain good attentions by only using the independent attention learning for some samples with attributes ``bangs" and ``receding hairline". 
In fact, attributes such as ``receding hairline" are usually treated as typical hard-cases for recognition. It is naturally difficult to capture their part-based features as their spatially discriminative regions have large variations for different samples, which further results in the difficulty of prediction.
However, we see that GAL can produce well-covered group attentions to capture the hairline region, which can be leveraged for all facial attributes around the hairline
and overcome the hard-case attribute problem.

\section{Group Affinity Matrix}
\begin{figure}[h!]
\begin{center}
   \includegraphics[width=0.8\linewidth]{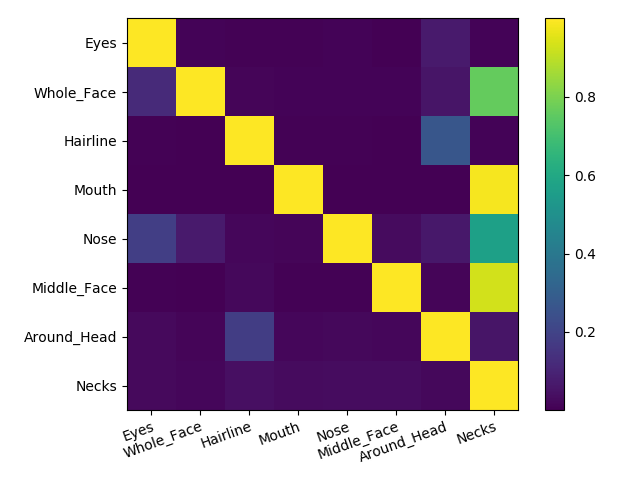}
\end{center}
   \caption{Group affinity matrix learned from LFWA. Each grid shows the effect from the part-based groups on the x-axis to the part-based groups on the y-axis. The affinity value is scaled from 0 (smallest) to 1 (largest).}
\label{fig:graphvis}
\end{figure}

We also visualize the learned correlations between part-based groups from GCL. Fig.~\ref{fig:graphvis} shows the estimated group affinity matrix learned on LFWA. This matrix demonstrates how the groups on the x-axis affect the groups on the y-axis. For instance, it shows that the ``neck" group has significant effects on the ``middle face" and ``mouth" groups than others. Another important discovery is that we can see the affinity matrix is not symmetric, where most diagonal values are non-identical. This observation indicates most correlations of groups are directed, which further supports our claim.

\section{Backbone Setting}
In the ablation study, we provide the results for our baseline, the network only using GAL and our MGG-Net with different backbone networks. In this section, we will specify our settings for this study.

For AlexNet, we select 3rd layer and 4th layer to apply our approach. The sizes of input features are $27\times27\times192$ and $13\times13\times384$ respectively. 

For VGG16, we select the layer Conv3-3 and the layer Conv4-3 to apply our approach. The sizes of input features  are $28\times28\times256$ and $14\times14\times512$ respectively. 

For ResNet34, we select the last layer of block 3 and the last layer of block 4 to apply our approach. The sizes of input features are $28\times28\times128$ and $14\times14\times256$ respectively.

\section{Analysis of Design Choice} Lastly, we show results with different designs in Table~\ref{Tab:gcnc}. It shows that introducing more blocks to apply our method can further improve the performance. Yet, the 3-blocks design will also bring more parameters to our model. Therefore, we only utilize 2 blocks to achieve a fair trade-off between accuracy and complexity.

\begin{table}[h!]
  \begin{center}
    \resizebox{0.4\textwidth}{!}{
    \begin{tabular}{c|c|c} % <-- Alignments: 1st column left, 2nd middle and 3rd right, with vertical lines in between
      \textbf{Design} & \textbf{Accuracy (\%)} & \textbf{\#Parameter (M)} \\
    %   \textbf{bone} & \textbf{(\%)} & \textbf{(\%)} & \textbf{Graph (\%)} \\
      \hline
      \hline
      Baseline  & 85.63 & 11 \\ \hline
      Block 3 & 86.67 & 11.6  \\ \hline
      Block 3+4 & 87.20 & 12.8  \\ \hline
      Block 3+4+2 & 87.27 & 13.2  \\ \hline
      Block 3+4+5 & 87.31 & 17.5  \\ \hline
    \end{tabular}}
  \end{center}
  \caption{\label{Tab:gcnc} Mean prediction accuracy (in \%) on LFWA and parameter numbers (in M) of MGG-Net with different designs.} 
\end{table}

\bibliographystyle{IEEEbib}
\bibliography{icme2020}